\documentclass[11pt,a4paper]{article}
\usepackage[hyperref]{acl2021}
\usepackage{times}
\usepackage{latexsym}

\usepackage{microtype}
\usepackage{graphicx}
\usepackage{booktabs}  
\usepackage{multirow}
\usepackage{subfigure}
\usepackage{verbatim}
\usepackage{CJKutf8} 
\usepackage{flushend}
\usepackage{makecell}
\aclfinalcopy 

\title{Hi-Transformer: Hierarchical Interactive Transformer for Efficient and Effective Long Document Modeling}

\author{Chuhan Wu$^\dagger$~~~~Fangzhao Wu$^\ddagger$~~~~Tao Qi$^\dagger$~~~~Yongfeng Huang$^\dagger$\\
    $^\dagger$Department of Electronic Engineering \& BNRist, Tsinghua University, Beijing 100084, China  \\
     $^\ddagger$Microsoft Research Asia, Beijing 100080, China\\
  {\tt\{wuchuhan15, wufangzhao, taoqi.qt\}@gmail.com} \\ {\tt yfhuang@tsinghua.edu.cn}
  }
\date{}

\begin{document}
\maketitle

\begin{abstract}

Transformer is important for text modeling.
However, it has difficulty in handling long documents due to the quadratic complexity with input text length.
In order to handle this problem, we propose a hierarchical interactive Transformer (Hi-Transformer) for efficient and effective long document modeling.
Hi-Transformer models documents in a hierarchical way, i.e., first learns sentence representations and then learns document representations.
It can effectively reduce the complexity and meanwhile capture global document context in the modeling of each sentence.
More specifically, we first use a sentence Transformer to learn the representations of each sentence.
Then we use a document Transformer to model the global document context from these sentence representations.
Next, we use another sentence Transformer to enhance sentence modeling using the global document context.
Finally, we use hierarchical pooling method to obtain document embedding.
Extensive experiments on three benchmark datasets validate the efficiency and effectiveness of Hi-Transformer in long document modeling.

\end{abstract}

\section{Introduction}

Transformer~\cite{vaswani2017attention} is an effective architecture for text modeling, and has been an essential component in many state-of-the-art NLP models like BERT~\cite{devlin2019bert,radford2019language,yang2019xlnet,wu2021da}.
The standard Transformer needs to compute a dense self-attention matrix based on the interactions between each pair of tokens in text, where the computational complexity is proportional to the square of text length~\cite{vaswani2017attention,wu2020improving}.
Thus, it is difficult for Transformer to model long documents efficiently~\cite{child2019generating}.

There are several methods to accelerate Transformer for long document modeling~\cite{wu2019lite,kitaev2019reformer,wang2020linformer,qiu2020blockwise}.
One direction is using Transformer in a hierarchical manner to reduce sequence length, e.g., first learn sentence representations and then learn document representations from sentence representations~\cite{zhang2019hibert,yang2020beyond}.
However, the modeling of sentences is agnostic to the global document context, which may be suboptimal because the local context within sentence is usually insufficient.
Another direction is using a sparse self-attention matrix instead of a dense one.
For example, \citet{beltagy2020longformer} proposed to combine local self-attention with a dilated sliding window and sparse global attention.
\citet{zaheer2020big} proposed to incorporate a random sparse attention mechanism to model the interactions between a random set of tokens.
However, these methods cannot fully model the global context of document~\cite{tay2020efficient}.

In this paper, we propose a hierarchical interactive Transformer (\textit{Hi-Transformer})\footnote{https://github.com/wuch15/HiTransformer.} for efficient and effective long document modeling, which models documents in a hierarchical way to effectively reduce the complexity and at the same time can capture the global document context for sentence modeling. 
In \textit{Hi-Transformer}, we first use a sentence Transformer to learn the representation of each sentence within a document.
Next, we use a document Transformer to model the global document context from these sentence representations.
Then, we use another sentence Transformer to further improve the modeling of each sentence with the help of the global document context.
Finally, we use hierarchical pooling method to obtain the document representation.
Extensive experiments are conducted on three benchmark datasets. The results show that \textit{Hi-Transformer} is both efficient and effective in long document modeling.

\begin{figure*}[!t]
  \centering 
      \includegraphics[width=0.99\linewidth]{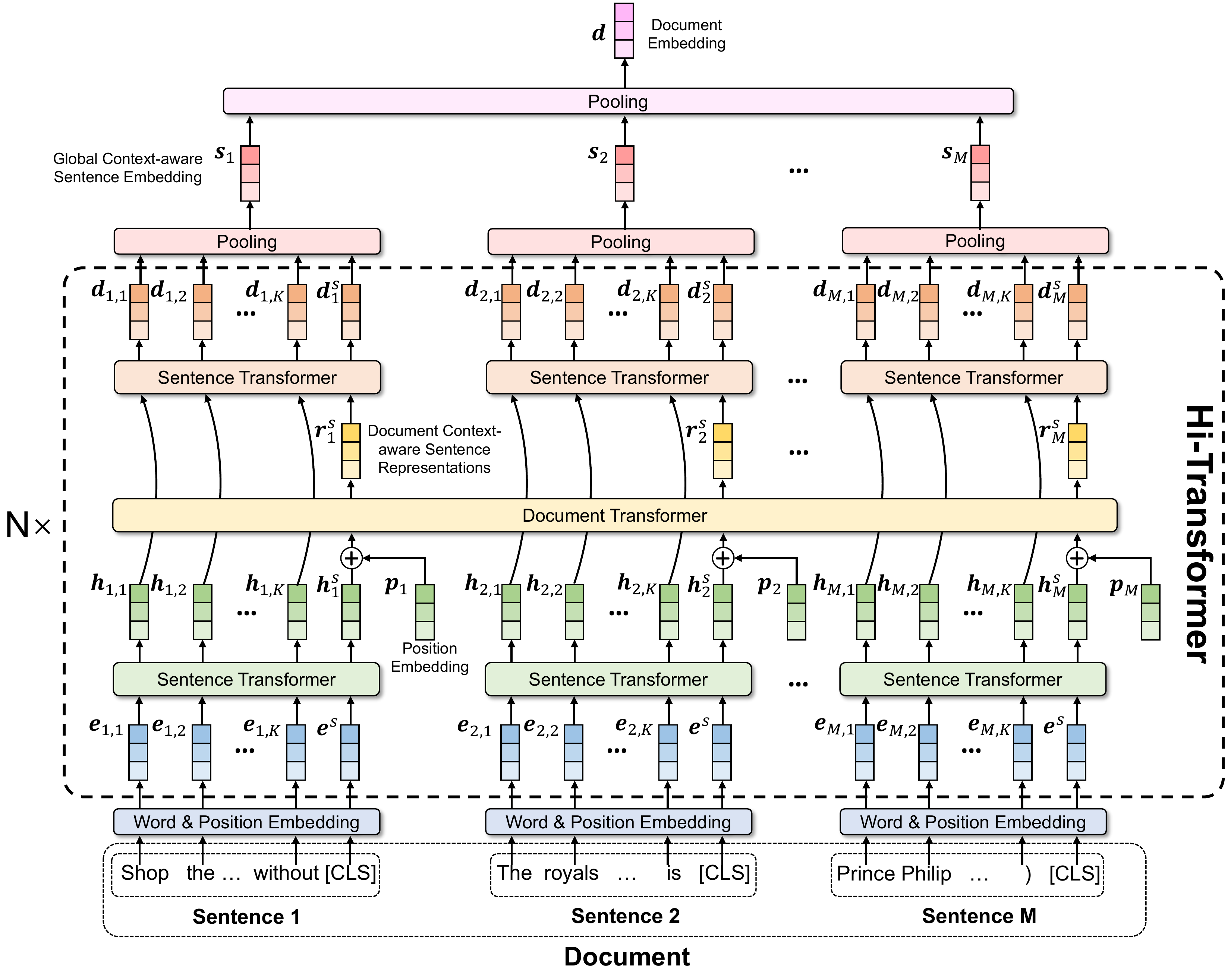}
  \caption{The architecture of \textit{Hi-Transformer}.}\label{fig.model}

\end{figure*}

\section{Hi-Transformer}\label{sec:Model}

In this section, we introduce our hierarchical interactive Transformer (\textit{Hi-Transformer}) approach for efficient and effective long document modeling.
Its framework is shown in Fig.~\ref{fig.model}.
It uses a hierarchical architecture that first models the contexts within a sentence, next models the document contexts by capturing the interactions between sentences, then employs the global document contexts to enhance sentence  modeling, and finally uses hierarchical pooling techniques to obtain document embeddings.
In this way, the input sequence length of each Transformer is much shorter than directly taking the word sequence in document as input, and the global contexts can be fully modeled.
The details of \textit{Hi-Transformer} are introduced as follows.

\subsection{Model Architecture}

\textit{Hi-Transformer} mainly contains three modules, i.e., sentence context modeling, document context modeling and global document context-enhanced sentence modeling.
The sentence-level context is first modeled by a sentence Transformer.
Assume a document contains $M$ sentences, and the words in the $i$-th sentence are denoted as $[w_{i,1}, w_{i,2}, ..., w_{i,K}]$ ($K$ is the sentence length).
We insert a ``[CLS]'' token (denoted as $w^s$) after the end of each sentence.
This token is used to convey the contextual information within this sentence.
The sequence of words in each sentence is first converted into a word embedding sequence via a word and position embedding layer.
Denote the word embedding sequence for the $i$-th sentence as $[\mathbf{e}_{i,1}, \mathbf{e}_{i,2}, ..., \mathbf{e}_{i,K},\mathbf{e}^s]$.
Since sentence length is usually short, we apply a sentence Transformer to each sentence to fully model the interactions between the words within this sentence.
It takes the word embedding sequence as the input, and outputs the contextual representations of words, which are denoted as $[\mathbf{h}_{i,1}, \mathbf{h}_{i,2}, ..., \mathbf{h}_{i,K}, \mathbf{h}^s_i]$.
Specially, the representation $\mathbf{h}^s_i$ of the ``[CLS]'' token is regarded as the sentence representation.

Next, the document-level context is modeled by a document Transformer from the representations of the sentences within this document.
Denote the embedding sequence of sentences in this document as $[\mathbf{h}_1^s, \mathbf{h}_2^s, ..., \mathbf{h}_M^s]$.
We add a sentence position embedding (denoted as $\mathbf{p}_i$ for the $i$-th sentence) to the sentence representations to capture sentence orders.
We then apply a document Transformer to these sentence representations to capture the global context of document, and further learn document context-aware sentence representations, which are denoted as $[\mathbf{r}_1^s, \mathbf{r}_2^s, ..., \mathbf{r}_M^s]$.

Then, we use the document context-aware sentence representations to further improve the sentence context modeling by propagating the global document context to each sentence.
Motivated by~\cite{guo2019star}, we apply another sentence Transformer to the hidden word representations and the document-aware sentence representation for each sentence.
It outputs a document context-aware word representation sequence for each sentence, which is denoted as $[\mathbf{d}_{i,1}, \mathbf{d}_{i,2}, ..., \mathbf{d}_{i,K}, \mathbf{d}^s_i]$.
In this way, the contextual representations of words can benefit from both local sentence context and global document context.

By stacking multiple layers of  \textit{Hi-Transformer}, the contexts within a document can be fully modeled.
Finally, we use hierarchical pooling~\cite{wu2020attentive} techniques to obtain the document embedding.
We first aggregate the document context-aware word representations in each sentence into a global context-aware sentence embedding $\mathbf{s}_i$, and then aggregate the global context-aware embeddings of sentence within a document into a unified document embedding $\mathbf{d}$, which is further used  for downstream tasks.

\subsection{Efficiency Analysis}

In this section, we provide some discussions on the computational complexity of \textit{Hi-Transformer}.
In sentence context modeling and document context propagation, the total computational complexity is $O(M\cdot K^2\cdot d)$, where $M$ is sentence number with a document, $K$ is sentence length, and $d$ is the hidden dimension.
In document context modeling, the computational complexity is $O(M^2\cdot d)$.
Thus, the total computational cost is $O(M\cdot K^2\cdot d+M^2\cdot d)$.\footnote{Note that \textit{Hi-Transformer} can be combined with other existing techniques of efficient Transformer to further improve the efficiency for long document modeling.}
Compared with the standard Transformer whose computational complexity is $O(M^2\cdot K^2\cdot d)$, \textit{Hi-Transformer} is much more efficient.
\section{Experiments}\label{sec:Experiments}

\begin{table*}[!t]
\centering
\begin{tabular}{ccccccc}
 \Xhline{1.5pt}
\multicolumn{1}{c}{\textbf{Dataset}} & \multicolumn{1}{c}{\textbf{\#Train}} & \multicolumn{1}{c}{\textbf{\#Val}} & \multicolumn{1}{c}{\textbf{\#Test}} & \multicolumn{1}{c}{\textbf{Avg. \#word}} & \multicolumn{1}{c}{\textbf{Avg. \#sent}} & \multicolumn{1}{l}{\textbf{\#Class}} \\ \hline
Amazon                               & 40.0k                                & 5.0k                               & 5.0k                                & 133.38                                   & 6.17                                     & 5                                    \\
IMDB                                 & 108.5k                               & 13.6k                              & 13.6k                               & 385.70                                   &  15.29                                     & 10                                   \\
MIND                                 & 128.8k                               & 16.1k                              & 16.1k                               & 505.46                                   & 25.14                                    & 18                                   \\  \Xhline{1.5pt}
\end{tabular}
\caption{Statistics of datasets.}\label{dataset}
\end{table*}
\subsection{Datasets and Experimental Settings}

Our experiments are conducted on three benchmark document modeling datasets.
The first one is Amazon Electronics~\cite{he2016ups} (denoted as Amazon), which is for product review rating prediction.\footnote{https://jmcauley.ucsd.edu/data/amazon/}
The second one is IMDB~\cite{diao2014jointly},  a widely used dataset for movie review rating prediction.\footnote{https://github.com/nihalb/JMARS}
The third one is the MIND dataset~\cite{wu2020mind}, which is a large-scale dataset for news intelligence.\footnote{https://msnews.github.io/}
We use the content based news topic classification task on this dataset.
The detailed dataset statistics are shown in Table~\ref{dataset}.

In our experiments, we use the 300-dimensional pre-trained Glove~\cite{pennington2014glove} embeddings for initializing word embeddings.
We use two \textit{Hi-Transformers} layers in our approach and two Transformer layers in other baseline methods.\footnote{We also tried more Transformer layers for baseline methods but do not observe significant performance improvement in our experiments.}
We use attentive pooling~\cite{yang2016hierarchical} to implement the hierarchical pooling module.
The hidden dimension is set to 256, i.e.,  8 self-attention heads in total and the output dimension of each head is 32.
Due to the limitation of GPU memory, the input sequence lengths of vanilla Transformer and its variants for long documents are 512 and 2048, respectively.
The dropout~\cite{srivastava2014dropout} ratio is 0.2.
The optimizer is Adam~\cite{kingma2014adam}, and the learning rate is 1e-4.
The maximum training epoch is 3.
The models are implemented using the Keras library with Tensorflow backend.
The GPU we used is GeForce GTX 1080 Ti with a memory of 11 GB.
We use accuracy and macro-F scores as the performance metrics.
We repeat each experiment 5 times and report both average results and standard deviations.

\begin{table*}[!t]
\resizebox{0.98\textwidth}{!}{
\begin{tabular}{lcccccc}
\Xhline{1.5pt}
\multicolumn{1}{c}{\multirow{2}{*}{Methods}} & \multicolumn{2}{c}{Amazon}      & \multicolumn{2}{c}{IMDB} & \multicolumn{2}{c}{MIND} \\ \cline{2-7} 
\multicolumn{1}{c}{}                         & Accuracy       & Macro-F        & Accuracy    & Macro-F    & Accuracy    & Macro-F    \\ \hline
Transformer                                  & 65.23$\pm$0.38 & 42.23$\pm$0.37 & 51.98$\pm$0.48       & 42.76$\pm$0.49     & 80.96$\pm$0.22       & 59.97$\pm$0.24      \\
Longformer                                   & 65.35$\pm$0.44 & 42.45$\pm$0.41 & 52.33$\pm$0.40       & 43.51$\pm$0.42     & 81.42$\pm$0.25       & 62.68$\pm$0.26      \\
BigBird                                     & 66.05$\pm$0.48 & 42.89$\pm$0.46 & 52.87$\pm$0.51       & 43.79$\pm$0.50     & 81.81$\pm$0.29       & 63.44$\pm$0.31      \\
HI-BERT                     & 66.56$\pm$0.32 & 42.65$\pm$0.34 & 52.96$\pm$0.46       & 43.84$\pm$0.46     & 81.89$\pm$0.23       & 63.63$\pm$0.20      \\ \hline
Hi-Transformer                               & 67.24$\pm$0.35 & 43.69$\pm$0.32 & 53.78$\pm$0.49       & 44.54$\pm$0.47     & 82.51$\pm$0.25       & 64.22$\pm$0.22      \\  \Xhline{1.5pt}
\end{tabular}
}
\caption{The results of different methods on different datasets. } \label{table.performance2} 
\end{table*}

\begin{table}[h]
\centering
\begin{tabular}{lc}
\Xhline{1.5pt}
\multicolumn{1}{c}{\textbf{Method}} & \multicolumn{1}{c}{\textbf{Complexity}} \\ \hline
Transformer                         & $O(M^2\cdot K^2 \cdot d)$                          \\
Longformer                          & $O(T\cdot M\cdot K \cdot d)$                                             \\
BigBird                             & $O(T\cdot M\cdot K \cdot d)$                                             \\
HI-BERT            & $O(M\cdot K^2\cdot d+M^2\cdot d)$                         \\
Hi-Transformer                      & $O(M\cdot K^2\cdot d+M^2\cdot d)$                         \\ \Xhline{1.5pt}
\end{tabular}
\caption{Complexity of different methods. $K$ is sentence length, $M$ is the number of sentences in a document, $T$ is the number of positions for sparse attention, and $d$ is the hidden dimension.}\label{complexity}
\end{table}

\subsection{Performance Evaluation}
We compare \textit{Hi-Transformer} with several baselines, including: 
(1) \textit{Transformer}~\cite{vaswani2017attention}, the vanilla Transformer architecture;
(2) \textit{Longformer}~\cite{beltagy2020longformer}, a variant of Transformer with local and global attention for long documents;
(3) \textit{BigBird}~\cite{zaheer2020big}, extending \textit{Longformer} with random attention;
(4) \textit{HI-BERT}~\cite{zhang2019hibert}, using  Transformers at both word and sentence levels.
The results of these methods on the three datasets are shown in Table~\ref{table.performance2}.
We find that Transformers designed for long documents like \textit{Hi-Transformer} and \textit{BigBird} outperform the vanilla Transformer.
This is because vanilla Transformer cannot handle long sequence due to the restriction of computation resources, and truncating the input sequence leads to the loss of much useful contextual information.
In addition, \textit{Hi-Transformer} and \textit{HI-BERT} outperform \textit{Longformer} and \textit{BigBird}.
This is because the sparse attention mechanism used in \textit{Longformer} and \textit{BigBird} cannot fully model the global contexts within a document.
Besides, \textit{Hi-Transformer} achieves the best performance, and the t-test results show the improvements over  baselines are significant.
This is because \textit{Hi-Transformer} can incorporate global document contexts to enhance sentence modeling.

We also compare the computational complexity of these methods in Table~\ref{complexity}.
The complexity of \textit{Hi-Transformer} is much less than the vanilla Transformer and is comparable with other Transformer variants designed for long documents.
These results indicate the efficiency and effectiveness of \textit{Hi-Transformer}.

\subsection{Model Effectiveness}

\begin{figure}[!t]
  \centering
  \subfigure[Amazon.]{
    \includegraphics[height=1.7in]{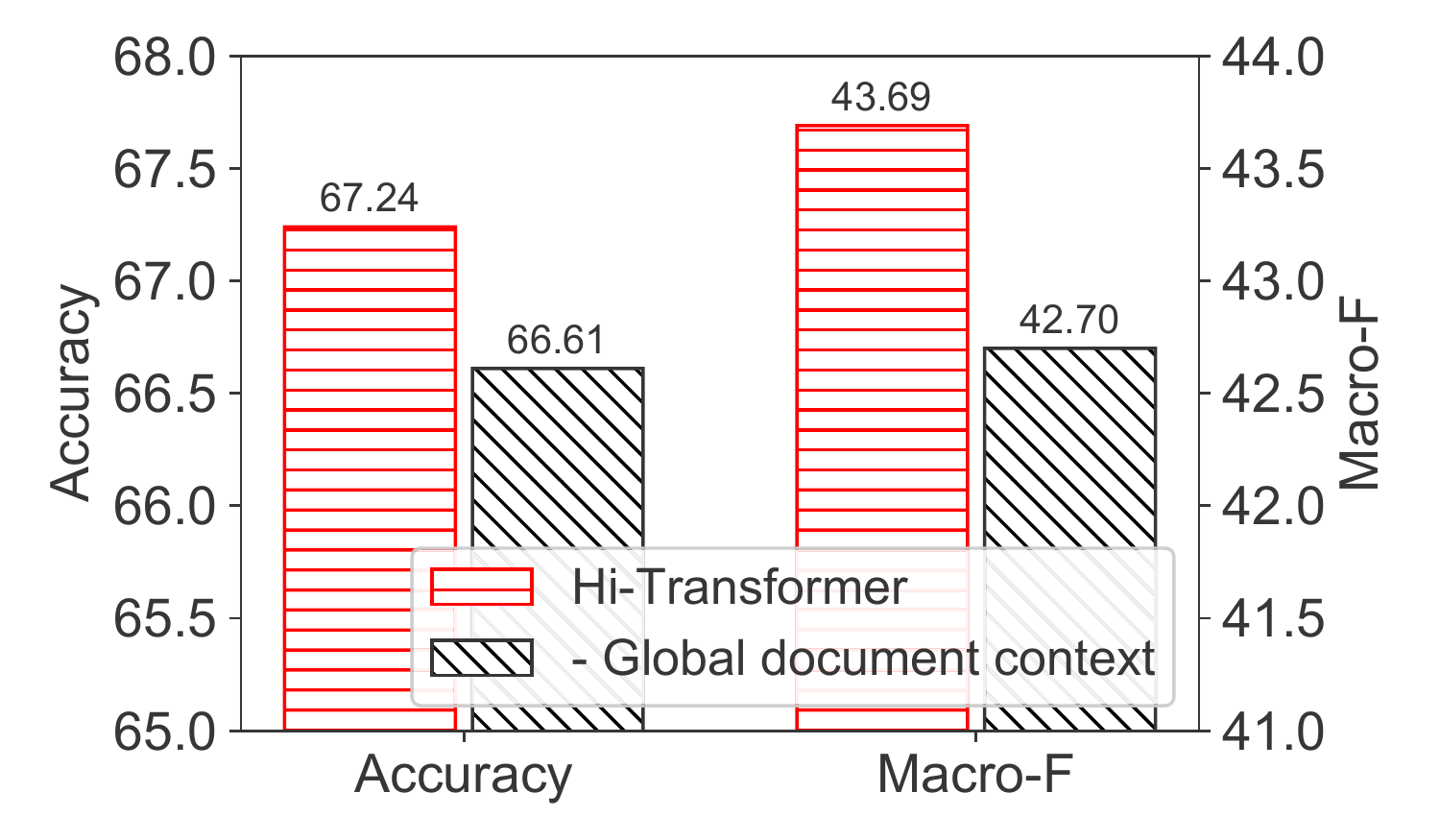}
  \label{fig.ab1}
  }
   \subfigure[IMDB.]{
      \includegraphics[height=1.7in]{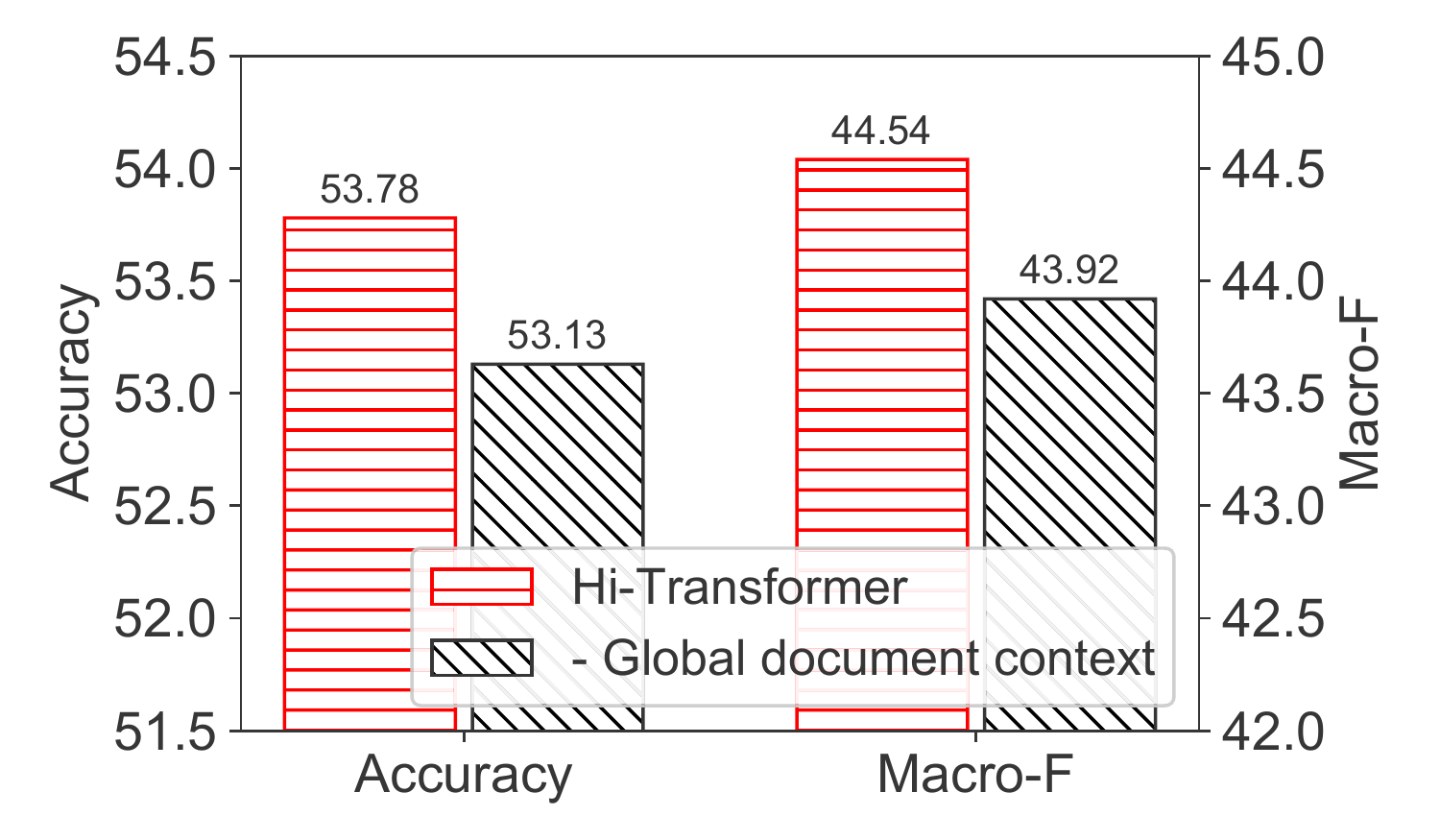}
  \label{fig.ab2}
  }
   \subfigure[MIND.]{
      \includegraphics[height=1.7in]{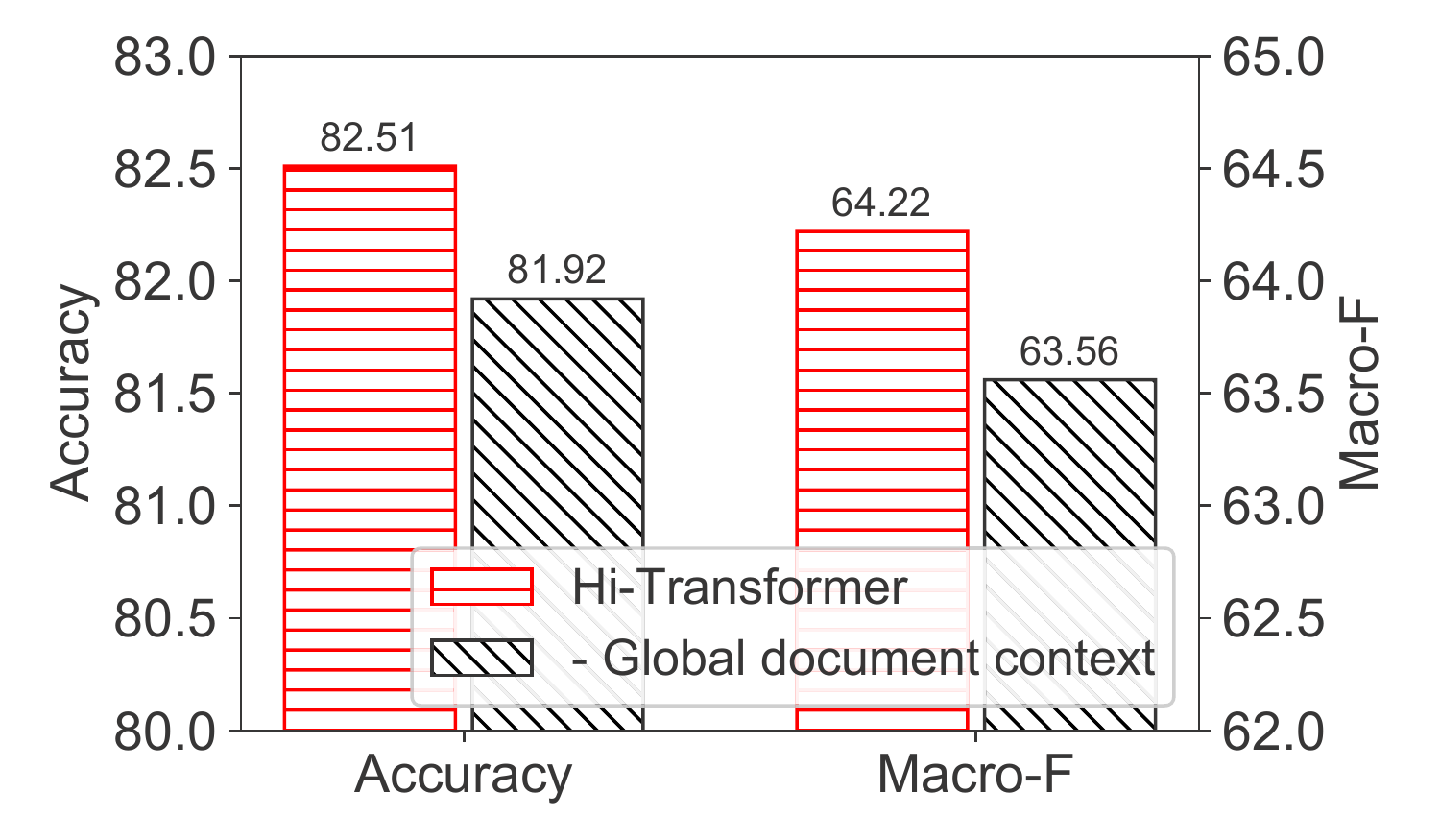}
  \label{fig.ab3}
  }
  \caption{Effectiveness of global document context propagation in \textit{Hi-Transformer} .}\label{fig.ab}

\end{figure}

Next, we verify the effectiveness of the global document contexts for enhancing sentence modeling in \textit{Hi-Transformer}.
We compare \textit{Hi-Transformer} and its variants without global document contexts in Fig.~\ref{fig.ab}.
We find the performance consistently declines when the global document contexts are not encoded into sentence representations.
This is because the local contexts within a single sentence may be insufficient for accurate sentence modeling, and global  contexts in the entire document can provide rich complementary information for sentence understanding.
Thus,  propagating the document contexts to enhance sentence modeling can improve long document modeling.

\subsection{Influence of Text Length}

\begin{figure}[!t]
  \centering
  \subfigure[Accuracy.]{
    \includegraphics[height=1.7in]{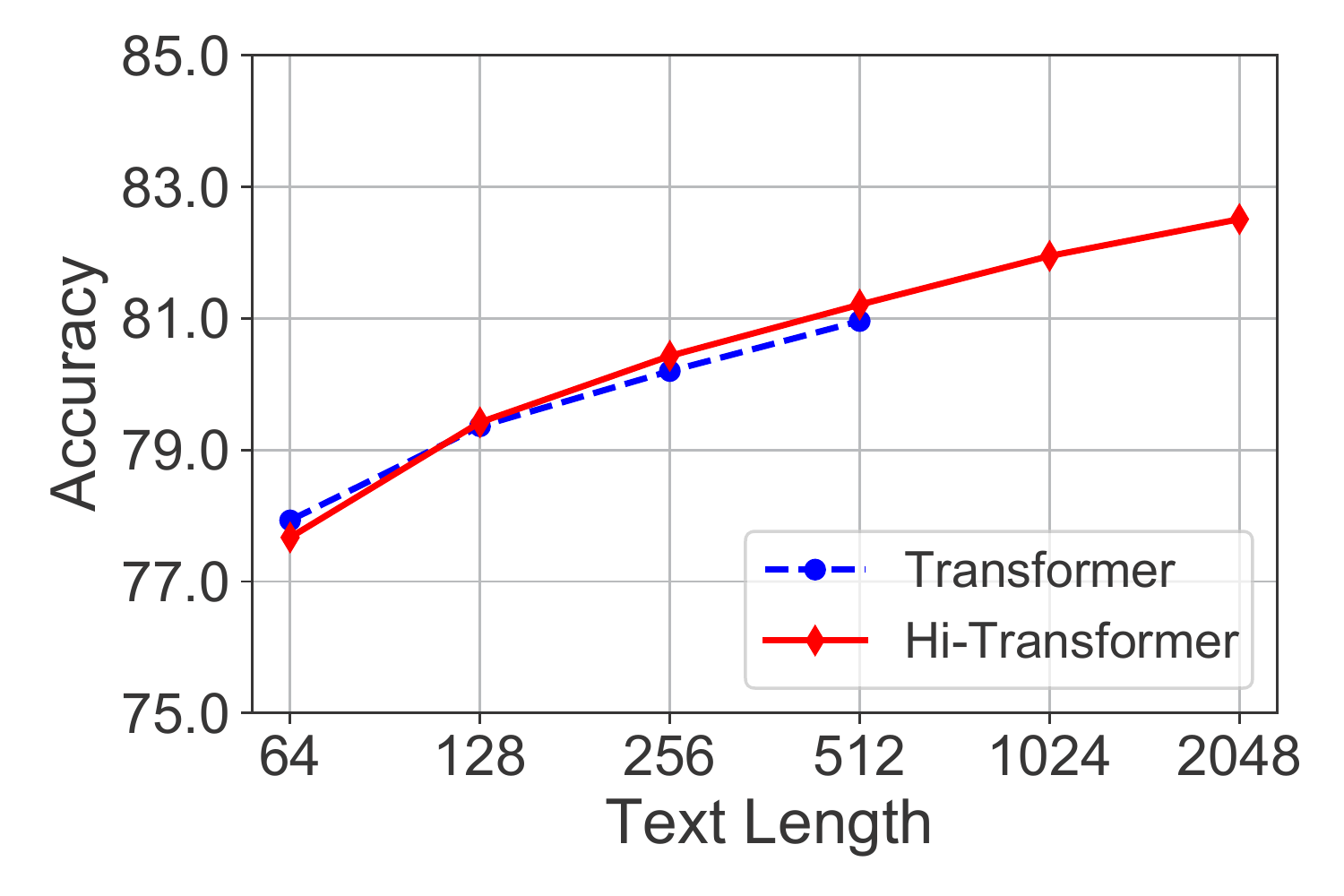}
  \label{fig.len1} 
  }
   \subfigure[Macro-F.]{
      \includegraphics[height=1.7in]{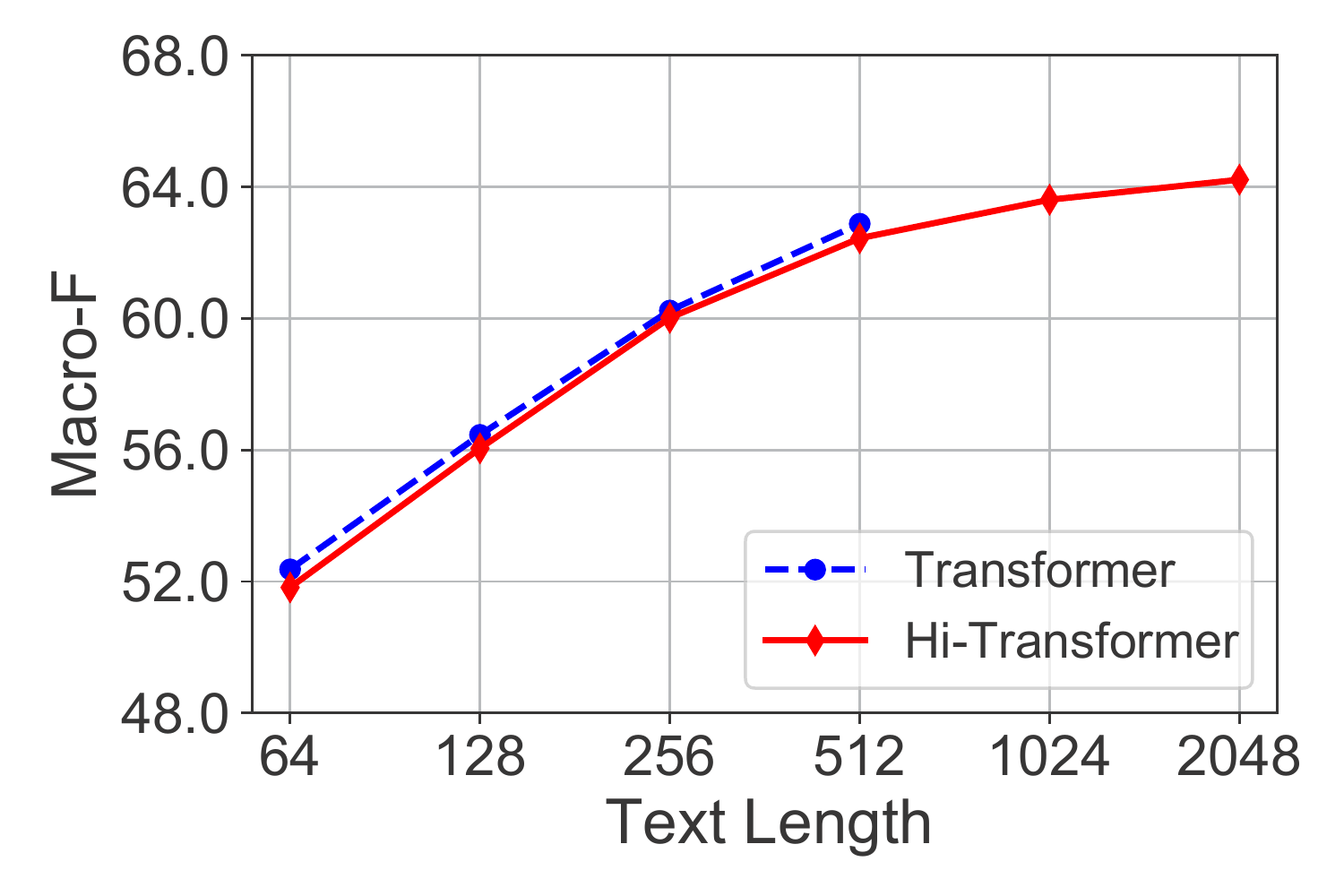}
  \label{fig.len2}
  } 
   \subfigure[Training time per layer.]{
      \includegraphics[height=1.7in]{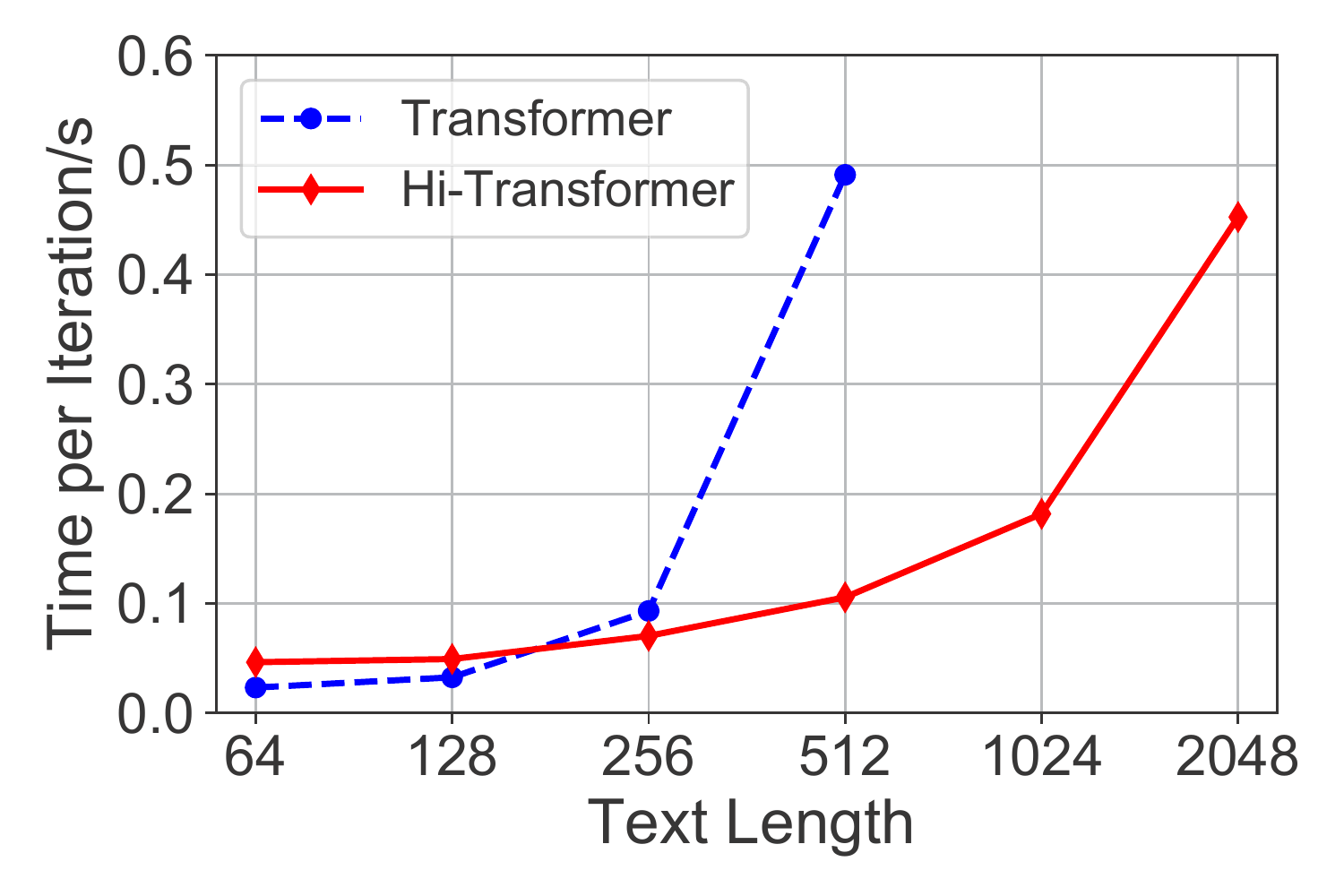}
  \label{fig.len3}
  }
  \caption{Influence of input text length on performance and training time on the MIND dataset.}\label{fig.len}
\end{figure}

Then, we study the influence of text length on the model performance and computational cost.
Since the documents in the MIND dataset are longest, we conduct experiments on MIND to compare the model performance as well as the training time per layer of \textit{Transformer} and \textit{Hi-Transformer} under different input text length\footnote{The maximum length of \textit{Transformer} is 512 due to GPU memory limitation.}, and the results are  shown in Fig.~\ref{fig.len}.
We find the performance of both methods improves when longer text sequences are used.
This is intuitive because more information can be incorporated when longer text is input to the model for document modeling.
However, the computational cost of \textit{Transformer} grows very fast, which limits its maximal input text length.
Different from \textit{Transformer}, \textit{Hi-Transformer} is much more efficient and meanwhile can achieve better performance with longer sequence length.
These results further verify the efficiency and effectiveness of \textit{Hi-Transformer} in long document modeling.

\section{Conclusion}\label{sec:Conclusion}

In this paper, we propose a \textit{Hi-Transformer} approach for both efficient and effective long document modeling.
It incorporates a hierarchical architecture that first learns sentence representations and then learns document representations.
It can effectively reduce the computational complexity and meanwhile be aware of the global document contexts in sentence modeling to help understand document content accurately.
Extensive experiments on three benchmark datasets validate the  efficiency and effectiveness of \textit{Hi-Transformer} in long document modeling.

\section*{Acknowledgments}
This work was supported by the National Natural Science Foundation of China under Grant numbers U1936216, U1936208, U1836204, and U1705261.
We are grateful to Xing Xie, Shaoyu Zhou, Dan Shen, and Zhisong Wang for their insightful comments and suggestions on this work.

\bibliographystyle{acl_natbib}
\bibliography{acl2021}
 

\end{document}